\documentclass{INTERSPEECH2023}

\newenvironment{tight}{%
\setlength{\abovedisplayskip}{3pt}
\setlength{\belowdisplayskip}{3pt}
}

\interspeechcameraready

\title{Accelerating Transducers through Adjacent Token Merging}
\name{Yuang Li$^{1 *}$\thanks{$^*$Work done during an internship at Microsoft.}, Yu Wu$^2$, Jinyu Li$^2$ Shujie Liu$^2$}

\address{
  $^1$University of Cambridge, $^2$Microsoft}
\email{yl807@eng.cam.ac.uk, \{Wu.Yu, jinyli, shujliu\}@microsoft.com}

\begin{document}

\maketitle

\begin{abstract}
Recent end-to-end automatic speech recognition (ASR) systems often utilize a Transformer-based acoustic encoder that generates embedding at a high frame rate. However, this design is inefficient, particularly for long speech signals due to the quadratic computation of self-attention.  To address this, we propose a new method, Adjacent Token Merging (A-ToMe), which gradually combines adjacent tokens with high similarity scores between their key values. In this way, the total time step could be reduced, and the inference of both the encoder and joint network is accelerated.  Experiments on LibriSpeech show that our method can reduce 57\% of tokens and improve the inference speed on GPU by 70\% without any notable loss of accuracy. Additionally, we demonstrate that A-ToMe is also an effective solution to reduce tokens in long-form ASR, where the input speech consists of multiple utterances.

\end{abstract}
\section{Introduction}

The area of end-to-end (E2E) automatic speech recognition (ASR) has seen significant progress in recent years \cite{miao2015eesen, las, watanabe2017hybrid, he2019streaming, Li2019RNNT, saon2021advancing, E2EOverview}, and three main approaches have emerged: Connectionist Temporal Classification (CTC)~\cite{graves2006connectionist}, Attention-based Encoder-Decoder (AED)~\cite{chorowski2015attention}, and Recurrent Neural Network Transducers (RNN-T)~\cite{graves2012sequence}. These methods differ in how they handle the alignment between speech and text tokens. AED uses cross-attention, while CTC and RNN-T use redundant symbols like ``blank''. 
The encoder of all these models processes fine-grained acoustic embedding at a high frame rate
, leading to high computational costs. Given that the frequency of acoustic tokens is much higher than that of text tokens, such as phonemes 
or word pieces
, significant redundancy exists. Hence, reducing the sequence length within the encoder is crucial for improving the efficiency of E2E ASR.

Adaptive subsampling techniques have been extensively researched in the field of Natural Language Processing, with token pruning being one of the most popular approaches~\cite{goyal2020power, kim2021length, wang2021spatten, kim2022learned}. Token pruning involves removing tokens with low importance scores, which are usually determined by the cumulative attention score in a multi-head attention mechanism. The amount of pruned tokens can be determined through a fixed configuration~\cite{wang2021spatten}, a learned threshold~\cite{kim2022learned}, or through evolutionary search~\cite{kim2021length}. These methods are often evaluated on sequence-level classification tasks rather than sequence-to-sequence tasks. For ASR, the majority of research focused on fixed-length subsampling such as progressively downsampling through convolutional layers~\cite{burchi2021efficient, huang2020conv}. Squeezeformer~\cite{kimsqueezeformer} further promoted the performance by using upsampling layer followed by downsampling. However, fixed-length subsampling can be sub-optimal as the duration of acoustic units varies considerably depending on the context and speaker. To address this issue, Meng et al.~\cite{meng2023compressing} proposed using a CIF~\cite{dong2020cif} module with the supervision of phoneme boundaries to achieve an adaptive rate in the Distill-Hubert~\cite{chang2022distilhubert}. Cuervo et al.~\cite{cuervovariable} proposed a two-level CPC network with a boundary predictor and an average pooling layer between the two levels.

In this study, we concentrate on a recently introduced adaptive subsampling technique called Token Merging~\cite{bolya2022tome}. The method was originally developed for use in Vision Transformers for classification tasks. It operates by merging tokens at any location that has high cosine similarity scores between their key values within the attention mechanism. However, it cannot be directly applied to the ASR task as preserving the temporal order of tokens is crucial. To address this issue, we propose a modified technique called Adjacent Token Merging (A-ToMe), which only merges tokens that are adjacent to each other. Furthermore, instead of merging a specific number of tokens, we introduce two different configurations to handle varying input lengths: fixed merge ratio and fixed merge threshold. Unlike previous studies, the proposed method does not explicitly predict boundaries. Instead, it gradually combines similar tokens to achieve a variable frame rate as the layers get deeper. 

Experiments were conducted on the LibriSpeech~\cite{panayotov2015librispeech} dataset using Transformer transducer~\cite{zhang2020transformer} as the baseline. We adjusted the number of merged tokens by changing the merge ratio or threshold. In most instances, when the total merge ratio was below 60\%, the model was able to maintain comparable word-error-rates (WERs) to the baseline while achieving a relative inference speedup of up to 35\% and 70\% on CPU and GPU respectively. Although the WER slightly increased as the number of merged tokens increased above 70\%, the performance remained significantly better than that of the convolutional subsampling. Furthermore, we extended our experiments to long-form ASR where history utterances are concatenated with current utterances to provide context information and showed that A-ToMe is even more crucial for accelerating the encoder when the input speech becomes longer. Finally, we found that the model trained with a fixed threshold can adapt to multiple thresholds during inference which can promote future research in the direction of on-demand token reduction.
\section{Methodology}

\subsection{Transformer transducer}

RNN-T~\cite{graves2012sequence} is composed of an encoder, a prediction network, and a joint network. The Transformer transducer~\cite{zhang2020transformer} extends the RNN-T by using a Transformer-based~\cite{vaswani2017attention} encoder that can effectively extract high-level representations $\mathbf{h}_t$ from acoustic features $\mathbf{X}$ (Equation~\ref{eq:1}). The prediction network generates embedding $\mathbf{z}_u$ based on previously predicted non-blank symbols $\mathbf{Y}_{<u}$ (Equation~\ref{eq:2}). The joint network, implemented as a feed-forward network (FFN), combines the output of the encoder and the prediction network, and its output is converted to token probabilities through a Softmax function (Equation~\ref{eq:3}).

 \begin{tight}
 \begin{align}
 \label{eq:1}
\mathbf{h}_t &= f_{enc}(\mathbf{X})\\
\label{eq:2}
\mathbf{z}_u &= f_{pred}(\mathbf{Y}_{<u})\\
\label{eq:3}
P(k|\mathbf{h}_t, \mathbf{z}_u) &= softmax(f_{joint}(\mathbf{h}_t, \mathbf{z}_u))
\end{align}
\end{tight}

The  ``blank'' token is used to help the alignment between the acoustic tokens from the encoder output with the text tokens. 
As there are many more acoustic tokens than text tokens, without token reduction, most output symbols are blank and will be removed in the final prediction.

\subsection{Adjacent token merging}


\begin{figure}[ht]
  \centering
  \includegraphics[width=\linewidth]{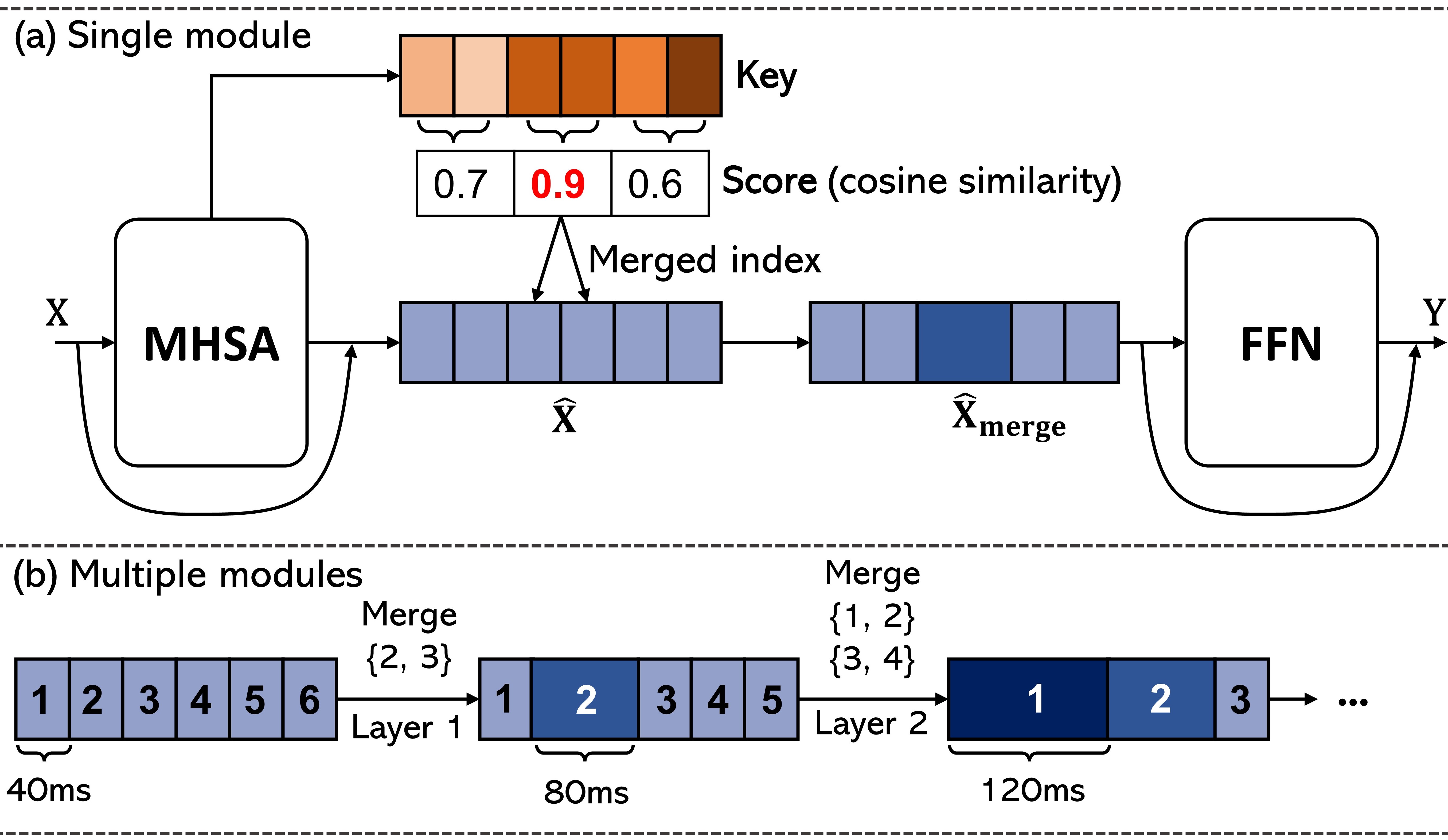}
  \caption{(a) A-ToMe module inside a Transformer layer. (b) The adaptive frame rate is achieved by stacking multiple modules.}
  \label{fig:token_merge}
\end{figure}

As shown in Figure~\ref{fig:token_merge} (a), the proposed A-ToMe module is inserted between the multi-head self-attention (MHSA) and FFN of a Transformer layer. This module utilizes the key values used in the self-attention calculation to determine the cosine similarity score between each pair of neighboring tokens. Tokens with high similarity scores are merged by taking their average, and the upper boundary for the merge ratio per layer is 50\%, which is equivalent to average pooling. Figure~\ref{fig:token_merge} (b) illustrates that multiple layers with the A-ToMe module can be stacked to achieve an adaptive frame rate, as merged tokens can be re-merged in subsequent layers. With $n$ modules and the original token length of $l$, the highest possible token length is $2^n \times l$. The A-ToMe is simple and efficient, requiring no additional parameters, and it can be implemented in parallel using PyTorch's~\cite{paszke2019pytorch} built-in functions without any loops.

To determine the number of tokens to merge, we employed two strategies that work for inputs of varying lengths: \textbf{1) Fixed merge threshold}: Tokens with a similarity score above a pre-defined threshold are merged. This strategy prevents dissimilar tokens from being merged at an earlier stage of the network, minimizing the loss of information. By adjusting the threshold, the number of merged tokens can be controlled, however, it is not possible to predict the exact number of merged tokens before inference. \textbf{2) Fixed merge ratio}: The similarity scores are ranked and a fixed ratio of tokens with the highest scores are merged. As the layer becomes deeper, the number of tokens decreases, leading to a corresponding decrease in the number of merged tokens. The advantage is that the number of output tokens can be pre-calculated based on the merge ratio. In Section~\ref{sec:short_form_results}, we demonstrate that the fixed merge ratio can also be interpreted as using a higher threshold for deeper layers.

\subsection{Long-form speech encoder}

ASR performance improves with longer sequences as more contextual information becomes available~\cite{narayanan2019recognizing, schwarz2020improving, hori2021advanced, masumura2021hierarchical}. In this paper, we adopted a simple approach to utilize historical utterances by concatenating the acoustic features from historical utterances $\{\mathbf{X}_{i-n},...,\mathbf{X}_{i-1}\}$ and the features $\mathbf{X}_i$ of the current utterance in order before the encoder (Equation~\ref{eq:4}). While the outputs contain both historical and current embeddings, the joint network only considers $\mathbf{H}_i$, the embedding corresponding to the current utterance.

 \begin{tight}
 \begin{align}
 \label{eq:4}
[\mathbf{H}_{i-n}; ...; \mathbf{H}_{i-1}; \mathbf{H}_i] = f_{enc}([\mathbf{X}_{i-n}; ...; \mathbf{X}_{i-1}; \mathbf{X}_i])
\end{align}
\end{tight}

\noindent This approach increases the computational intensity and memory consumption of the encoder due to the quadratic complexity of MHSA. Therefore, A-ToMe can be more important in this case. Additionally, different merging configurations can be applied to current and historical tokens, considering that current tokens may be more crucial. For instance, we can limit merging to only historical tokens or use a higher merge ratio for historical tokens than for current tokens.

\begin{table*}[t]
\small
\caption{The comparison between different merging configurations. The average of merged tokens, token length, and latency/speed is calculated based on the test-clean subset. The ratio of merged tokens and the average token length refer to the tokens after the encoder.}
\label{tab:short_form}
\centering
\begin{tabular}{c | c | c |c c | c c c c} 
\toprule
 & Merged & Token & \multicolumn{2}{c|}{Latency (s) / Speed} & \multicolumn{2}{c}{WER Dev} &\multicolumn{2}{c}{WER Test}\\
 Method & Tokens (\%) & Length (ms) & CPU & GPU & clean & other & clean & other\\
\midrule
 baseline & 0 & 40 & 3.66 / 1.00$\times$ & 1.07 / 1.00$\times$ & 2.57 & 5.79 & 2.79 & 6.01\\
subsampling$\times$2 & 50 & 80 & 2.99 / 1.22$\times$ & 0.67 / 1.59$\times$ & 2.71 & 6.08 & 2.90 & 6.29\\
subsampling$\times$4 & 75 & 160 & 2.16 / 1.70$\times$ & 0.50 / 2.16$\times$ & 3.07 & 6.77 & 3.15 & 6.86\\
 \midrule
 \multicolumn{9}{c}{A-ToMe (fixed merge ratio)}\\
 \midrule
ratio/layer=10\% & 46 & 74 & 2.86 / 1.28$\times$ & 0.74 / 1.46$\times$ & \textbf{2.63} & \textbf{5.86} & \textbf{2.79} & \textbf{5.90}\\
ratio/layer=15\% & 61 & 103 & 2.43 / 1.51$\times$ & 0.62 / 1.73$\times$ & 2.67 & 6.02 & 2.88 & 6.02\\
ratio/layer=20\% & 73 & 148 & 2.06 / 1.78$\times$ & 0.53 / 2.04$\times$ & 2.80 & 5.95 & 2.88 & 6.20\\
\midrule
\multicolumn{9}{c}{A-ToMe (fixed merge threshold)}\\
\midrule
threshold=0.90 & 42 & 69 & 3.09 / 1.18$\times$ & 0.78 / 1.37$\times$ & 2.79 & \textbf{5.74} & 2.90 & 6.17\\
threshold=0.85 & 57 & 93 & 2.70 / 1.35$\times$ & 0.63 / 1.70$\times$ & \textbf{2.66} & 5.78 & \textbf{2.89} & \textbf{5.96}\\
threshold=0.80 & 72 & 143 & 2.20 / 1.66$\times$ & 0.54 / 1.98$\times$ & 2.70 & 5.97 & 3.01 & 6.04\\
\bottomrule
\end{tabular}
\end{table*}

\section{Experiments}

\subsection{Experimental setup}

Evaluations were performed on the LibriSpeech dataset~\cite{panayotov2015librispeech}, which comprises 960 hours of speech. We report the WERs on dev-clean, dev-other, test-clean, and test-other subsets. Moreover, we measured the average inference latency per utterance for the test-clean subset on GPU and CPU. For GPU latency, we used beam search with a beam size of 16 on NVIDIA Tesla V100 32GB GPU. For CPU latency, we used a single core of Intel Xeon CPU E5-2673 and employed greedy search instead of beam search for efficiency. The WERs reported were obtained using the beam search decoding method.

The encoder of the Transformer transducer has a four-layer VGG-like convolutional network that reduces the frame rate by a factor of four, followed by 18 Transformer layers. Each Transformer layer consists of an MHSA with an attention dimension of 512 and eight heads, and an FFN with a hidden dimension of 2048. The encoder takes as input 80-dimensional filterbank features with a frame length of 25 ms and a stride of 10 ms. The prediction network is a two-layer LSTM~\cite{hochreiter1997long} with a hidden dimension of 1024, and the joint network has an embedding dimension of 512. 5000-wordpiece vocabulary~\cite{kudo2018sentencepiece} is used as the target. The whole model contains 94 million parameters, with the majority located in the encoder (59 million). We used a multitask loss function~\cite{jeon2021multitask} including RNN-T, AED, and CTC losses with weights of 1, 0.3, and 0.3 respectively. The model was trained from scratch for 400,000 steps with AdamW~\cite{loshchilovdecoupled} optimizer. Specaug~\cite{park2019specaugment} was adopted for better generalizations.



The Transformer encoder incorporates A-ToMe every three layers, specifically at layers 2, 5, 8, 11, 14, and 17. In addition to presenting results for the un-merged model, we also report the outcomes of a traditional subsampling technique, achieved by adding extra convolutional layers to the VGG-like downsampler, as a baseline comparison. In the long-form ASR experiments, only a small batch size can be used which leads to slow convergence if the transducer is trained from scratch. Hence, we fine-tuned the utterance-based models for 200,000 steps with history utterances. 



\subsection{Utterance-based ASR results}
\label{sec:short_form_results}




 \begin{figure}[h]
  \centering
  \includegraphics[width=0.96\linewidth]{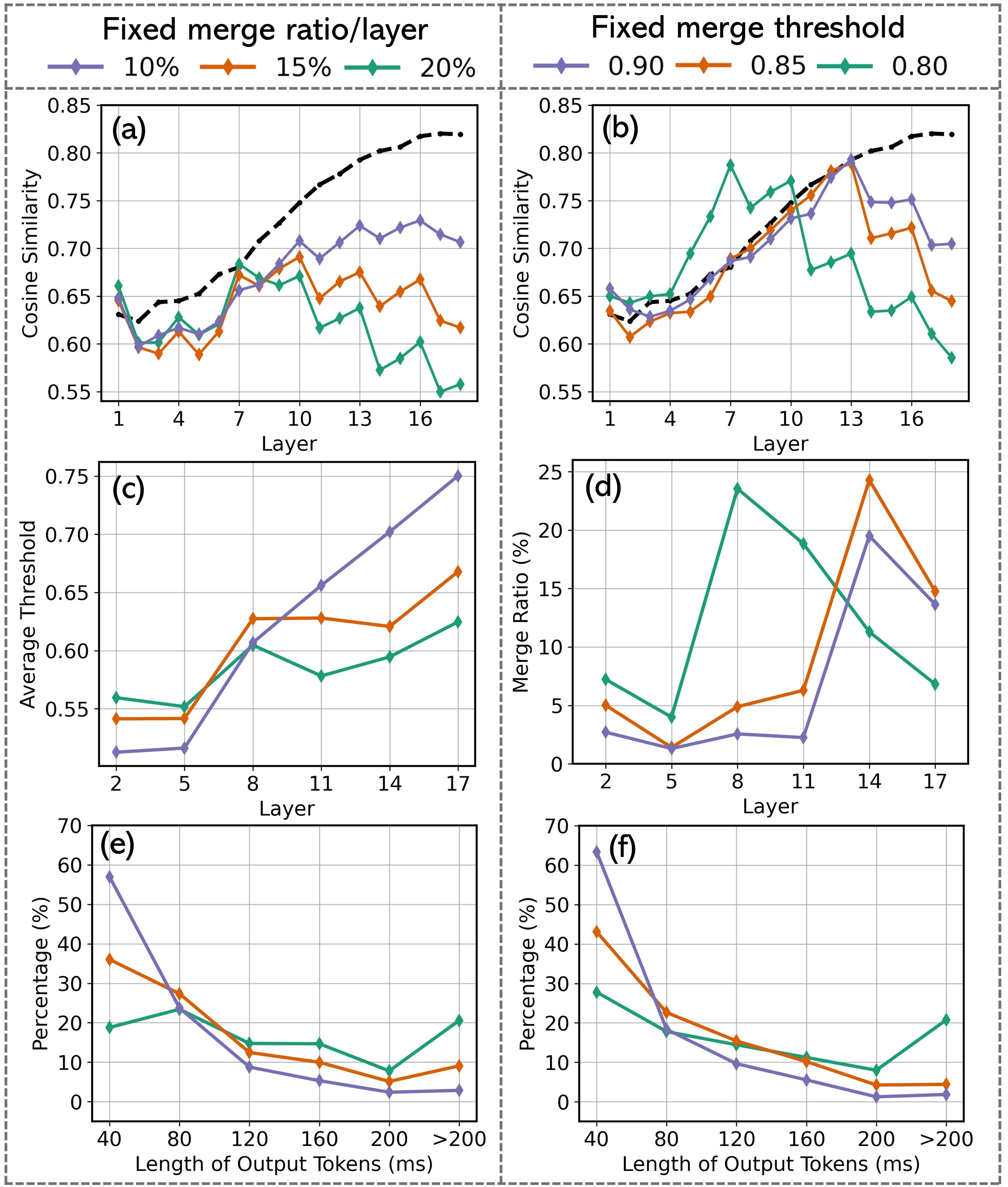}
  \caption{Visualizing different merge configurations for better understanding on the test-clean subset. (a, b) The change in cosine similarity between adjacent tokens from shallow to deep layers. (c) The average threshold at different layers when a fixed merge ratio is applied. (d) The average merge ratio at different layers when a fixed merge threshold is used. (e, f) Distribution of encoder output token lengths in percentages.}
  \label{fig:short_form}
\end{figure}

As shown in Table~\ref{tab:short_form}, the convolutional subsampling resulted in a significant increase in WERs, particularly on the dev-other and test-other subsets. For instance, when a subsampling rate of $\times2$ and $\times4$ was employed, there was a relative degradation in WER of 5\% and 14\% respectively, on the test-other subset. Fixed merge ratios led to a slight increase in WER as the number of merged tokens increased, but the impact was much smaller than that of convolutional subsampling. When the total merged tokens reached 73\% (comparable to subsampling$\times4$), the WER on test-other only increased by 3\% relative. Moreover, when the merge ratio per layer was 10\% and the total merged tokens were 46\%, there was no noticeable degradation compared to the baseline. In terms of speed, A-ToMe contributed to much lower E2E latencies compared to the baseline by accelerating the encoder and reducing forward steps in decoding. The speed on the CPU became 1.28 to 1.78 times faster when using merge ratios per layer between 10\% and 20\%. Evaluations on the CPU directly reflected the computations since operations were not paralleled. For GPU performance, A-ToMe was even more effective with a speed of 1.46 to 2.01 times faster since the bottleneck here is the iterative forward steps during decoding, which increases with the number of tokens and is hard to parallelize. Compared to fixed ratio merging, using fixed merge thresholds performed slightly better when the number of merged tokens was high. For example, when using a threshold of 0.85, approximately 57\% of tokens were merged with a negligible performance drop. However, the performance with the threshold of 0.9 was less satisfactory and the speedup on the CPU was more limited as fewer tokens at lower layers were merged.


Figure~\ref{fig:short_form} provides visualizations to better understand A-ToMe. As shown by the dashed line in Figure~\ref{fig:short_form} (a, b), without merging, the cosine similarity between adjacent tokens kept increasing as the layer became deeper, indicating considerable redundant information. With fixed merge ratios, the cosine similarity was kept at a relatively low level throughout the network, whereas with fixed merge thresholds, cosine similarity was reduced mostly at deep layers. This is because at low layers few tokens were merged as similarity scores were below the threshold. We can see from Figure~\ref{fig:short_form} (d) that with thresholds of 0.85 and 0.90, most tokens were merged at layers 14 and 17. For a lower threshold of 0.8, more tokens can be merged at shallower layers like layer 8. This also means that enforcing a fixed merge ratio is similar to using a lower threshold for shallow layers (Figure~\ref{fig:short_form} (c)). Figure~\ref{fig:short_form} (e, f) shows the adaptive token length achieved by A-ToMe. With aggressive merging configurations such as the threshold of 0.8, more than 20\% of tokens had lengths of more than 200 ms. For a lower merge ratio like 10\% per layer, more than 50\% of tokens remained unchanged at 40 ms. These visualizations highlight two main strengths of our approach: 1) variable token lengths instead of fixed lengths like 80 ms and 160 ms, and 2) gradual subsampling instead of subsampling all at once.

\subsection{Long-form ASR results}

\begin{table}[t]
\small
\caption{Long-form ASR utilizing A-ToMe. 'Merge history' represents the merging of only previous utterances, while 'Merge all' indicates the merging of both previous and current utterances.}
\label{tab:long_form}
\centering
\begin{tabular}{c | c | c c c c} 
\toprule
 &  & \multicolumn{2}{c}{WER Dev} &\multicolumn{2}{c}{WER Test}\\
 History & Merge & clean & other & clean & other\\
\midrule
0 & - & 2.57 & 5.79 & 2.79 & 6.01 \\
\midrule
1 & - & 2.54 & 5.45 & \textbf{2.61} & 5.64 \\
1 & history & 2.55 & \textbf{5.40} & 2.69 & 5.66 \\ 
1 & all & \textbf{2.48} & 5.47 & 2.74 & \textbf{5.61} \\
\midrule
2 & - & \textbf{2.35} & 5.20 & \textbf{2.57} & 
\textbf{5.38} \\
2 & history & 2.39 & \textbf{5.17} & 2.64 & 5.50 \\
2 & all & 2.42 & 5.31 & 2.67 & 5.49\\
\bottomrule
\end{tabular}
\end{table}

\begin{figure}[t]
  \centering
  \includegraphics[width=0.86\linewidth]{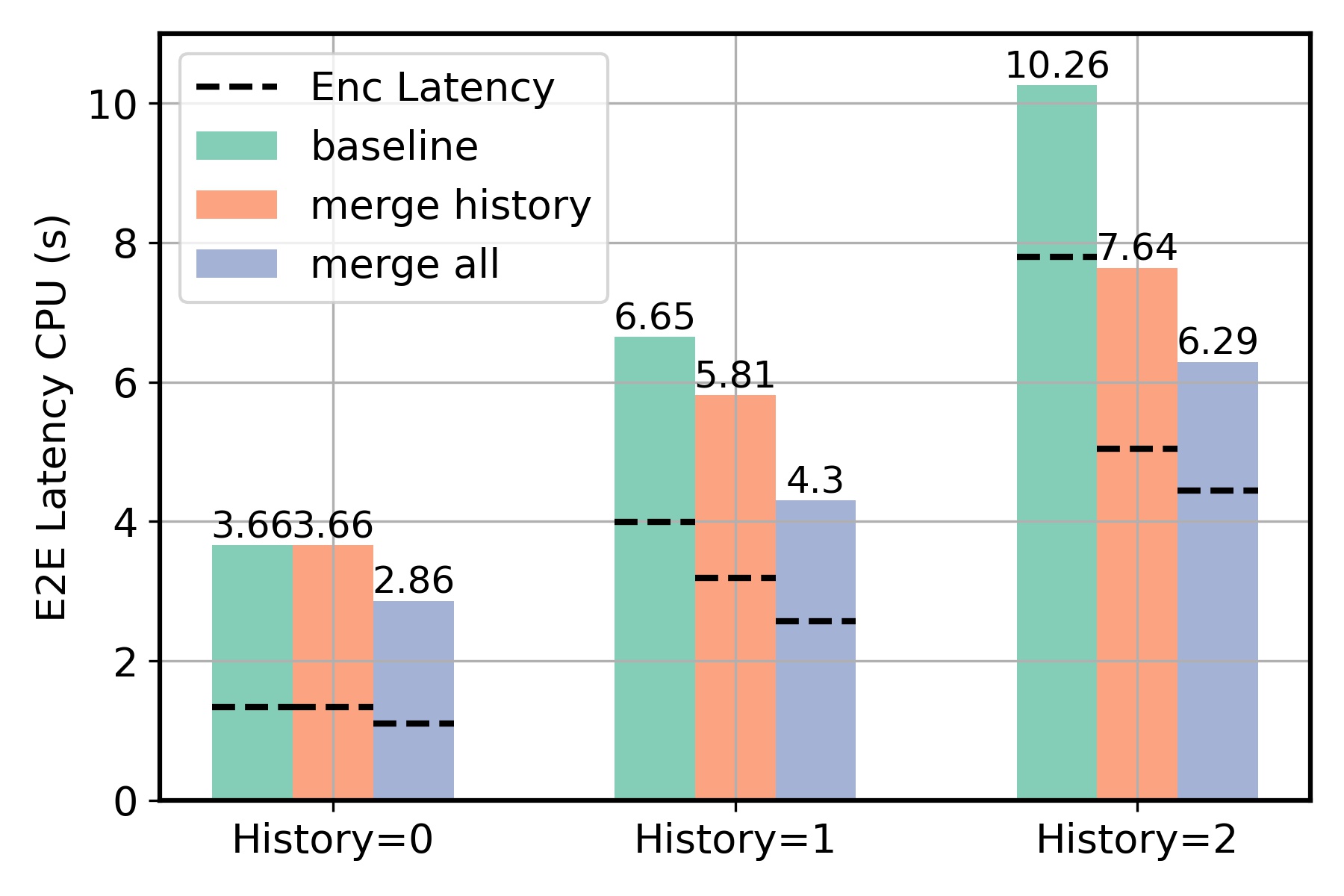}
  \caption{E2E latency on CPU with varying numbers of historical utterances. The region below the dashed line indicates the duration spent by the encoder, while the area above the dashed line represents the time consumed by the prediction and joint network during the decoding process.}
  \label{fig:speed}
\end{figure}

In the long-form ASR experiment, we investigated two configurations that significantly merge history tokens while preserving more current information. The first configuration involves merging only historical tokens, with a fixed merge ratio of 20\% per layer. The second configuration involves merging current tokens with a ratio of 10\% per layer, and historical tokens with a ratio of 20\% per layer. As shown in Table~\ref{tab:long_form}, ASR performance improved as more context was added. Without merging, the WER on test-other decreased from 6.01\% to 5.38\% when two historical utterances were used. When there was only one history utterance, the two merging configurations had similar WERs and were comparable to the results of the unmerged model. When there were two historical utterances, A-ToMe slightly affected the performance, and merging only historical tokens yielded slightly better results than merging both current and historical tokens. It is worth noting that using a merge ratio of 20\% per layer on historical tokens had a smaller impact on WERs than using it on current tokens. Figure~\ref{fig:speed} illustrates comparisons of E2E latency on the CPU. As the number of historical utterances increased from zero to two, there was a significant increase in latency from 3.66 seconds to 10.26 seconds when A-ToMe was not used. The encoder latency was primarily affected, whereas the rest of the model was less impacted since history tokens are removed after the encoder.  Furthermore, the speed gain from A-ToMe improves as sequences become longer, with a shift from primarily benefiting the computation after the encoder to benefiting the encoder itself.


\subsection{On-demand inference with different threshold}

\begin{figure}[ht]
  \centering
  \includegraphics[width=\linewidth]{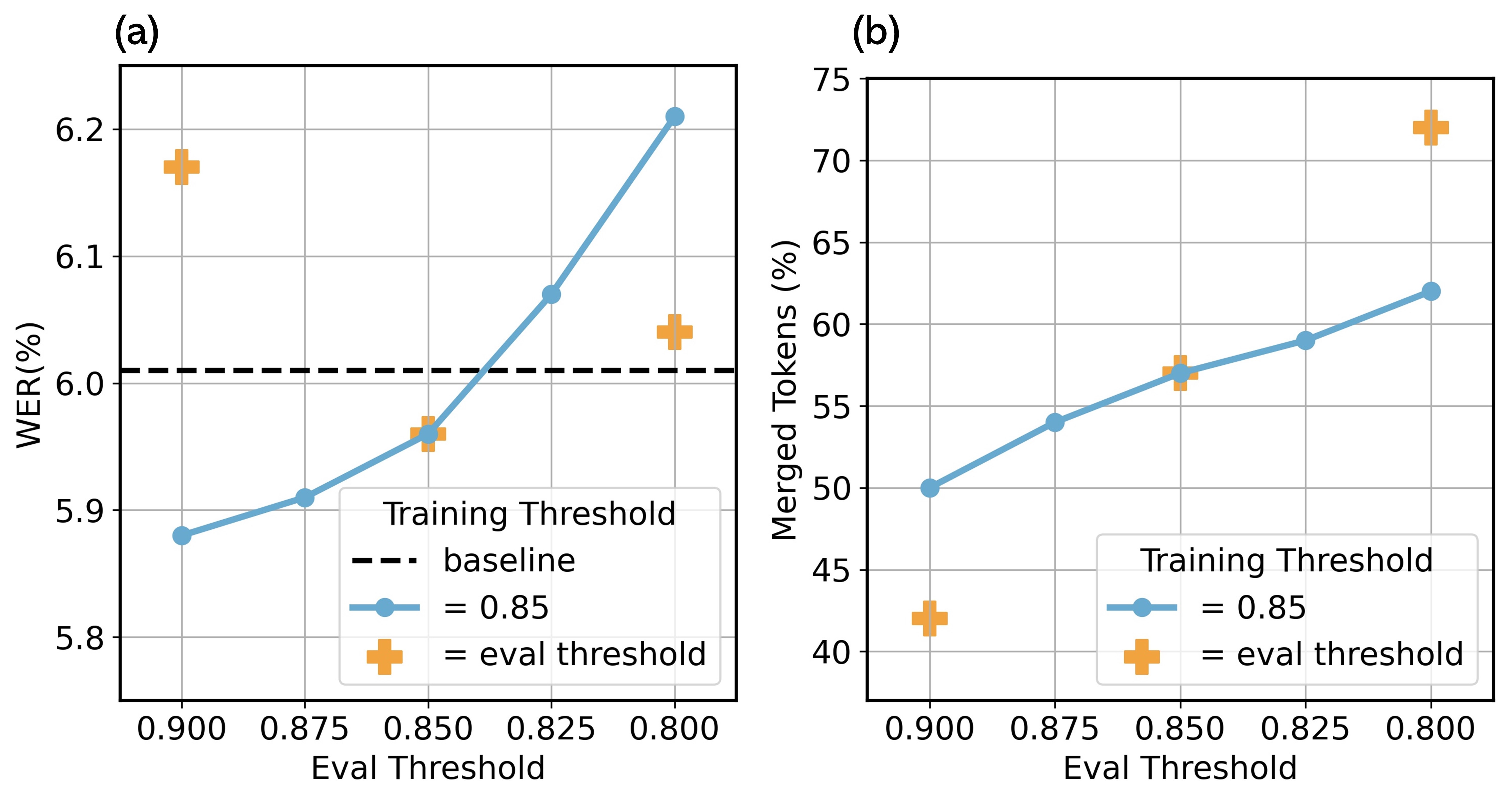}
  \caption{(a) The impact of applying distinct thresholds during inference compared to training on WER (test-other). (b) The proportion of merged tokens with different threshold setups.}
  \label{fig:adapt}
\end{figure}

On-demand compute reduction~\cite{cai2020once, vyas2022demand} involves training a model that can be adapted to various computational requirements at inference without retraining. We conducted preliminary experiments to examine the on-demand capability of A-ToMe. Figure~\ref{fig:adapt} (a) shows the WERs on test-other when the model was evaluated with different thresholds, even though it was trained with a fixed threshold of 0.85. Figure~\ref{fig:adapt} (b) illustrates the percentage of merged tokens while the threshold was adjusted. By modifying the threshold, we can control the number of merged tokens during inference while maintaining good performance, especially at high thresholds. However, we observed that the performance was not as good with low thresholds, such as 0.8. Additionally, when using an on-demand setup with thresholds between 0.8 and 0.9, the percentage of merged tokens had a narrower range than the traditional setup where the same threshold was used for training and evaluation.

\section{Conclusion}

In this paper, we proposed a novel adaptive subsampling method called Adjacent Token Merging that progressively reduces the number of tokens in the encoder of the Transformer transducer. We emphasized the importance of variable frame rates and gradual subsampling. Experiments on utterance-based and long-form ASR showed that our approach could accelerate inference substantially while having minimal impact on recognition performance. Additionally, our approach can provide more flexibility in designing efficient ASR models and on-demand neural networks, which will facilitate future research. Moving forward, we plan to investigate more sophisticated merging strategies, and we will adapt our approach for streaming ASR.

\clearpage
\bibliographystyle{IEEEtran}
\bibliography{mybib}

\end{document}